\documentclass[10pt,twocolumn,letterpaper]{article}

\usepackage{iccv}
\usepackage{times}
\usepackage{epsfig}
\usepackage{graphicx}
\usepackage{amsmath}
\usepackage{amssymb}
\usepackage{booktabs}
\usepackage{microtype}
\usepackage[table,xcdraw]{xcolor}
\usepackage{caption}
\usepackage{float}
\usepackage{placeins}
\usepackage{color, colortbl}
\usepackage{stfloats}
\usepackage{enumitem}
\usepackage{tabularx}
\usepackage{xstring}
\usepackage{multirow}
\usepackage{xspace}
\usepackage{url}
\usepackage{subcaption}
\usepackage{xcolor}
\usepackage[hang,flushmargin]{footmisc}
\usepackage{bbding}
\usepackage{multicol}
\usepackage[table]{xcolor}
\usepackage{algorithm}
\usepackage{algorithmic}
\usepackage{amsmath}
\definecolor{citecolor}{HTML}{0071bc}

\usepackage[breaklinks=true,bookmarks=false]{hyperref}

\iccvfinalcopy 

\def\iccvPaperID{4968}

\ificcvfinal\fi

\begin{document}

\title{Video Background Music Generation: Dataset, Method and Evaluation}

\author{
    Le Zhuo$^1$\thanks{Equal contribution.} \qquad
    Zhaokai Wang$^1$\footnotemark[1] \qquad
    Baisen Wang$^1$\footnotemark[1] \qquad
    Yue Liao$^1$\thanks{Corresponding author.} \qquad
    Chenxi Bao$^{1,2}$ \qquad
    \\
    Stanley Peng$^1$ \qquad
    Songhao Han$^1$ \qquad
    Aixi Zhang$^3$ \qquad 
    Fei Fang$^3$ \qquad 
    Si Liu$^1$ 
    \\
    $^1$Beihang University \quad 
    $^2$Edinburgh College of Art, University of Edinburgh \quad
    $^3$Alibaba Group
    \\
    {\tt\small zhuole1025@gmail.com, \{wzk1015, wbs2788\}@buaa.edu.cn, liaoyue.ai@gmail.com}
}

\maketitle
\ificcvfinal\thispagestyle{empty}\fi

\begin{abstract}

Music is essential when editing videos, but selecting music manually is difficult and time-consuming. Thus, we seek to automatically generate background music tracks given video input. This is a challenging task since it requires music-video datasets, efficient architectures for video-to-music generation, and reasonable metrics, none of which currently exist. To close this gap, we introduce a complete recipe including dataset, benchmark model, and evaluation metric for video background music generation. We present SymMV, a video and symbolic music dataset with various musical annotations. To the best of our knowledge, it is the first video-music dataset with rich musical annotations. We also propose a benchmark video background music generation framework named V-MusProd, which utilizes music priors of chords, melody, and accompaniment along with video-music relations of semantic, color, and motion features. To address the lack of objective metrics for video-music correspondence, we design a retrieval-based metric VMCP built upon a powerful video-music representation learning model. Experiments show that with our dataset, V-MusProd outperforms the state-of-the-art method in both music quality and correspondence with videos. We believe our dataset, benchmark model, and evaluation metric will boost the development of video background music generation. Our dataset and code are available at \textcolor{citecolor}{\url{https://github.com/zhuole1025/SymMV}}.

\end{abstract}

\vspace{-1mm}
\section{Introduction}
\vspace{-1mm}

Music plays a crucial role when creating videos, improving the overall quality of videos and enhancing the immersion for viewers. With the rapid growth of social platforms, the needs to find suitable music for videos extend from professional fields,~\emph{e.g.}, soundtrack production in the film industry, to amateur usages like video blogs and TikTok short videos. However, finding proper music for videos and making alignments are difficult and may even bring copyright issues. Thus, automatically generating background music for videos is of great value to a wide range of creators. 

\begin{figure*}[t]
    \centering
    \includegraphics[width=0.9\textwidth]{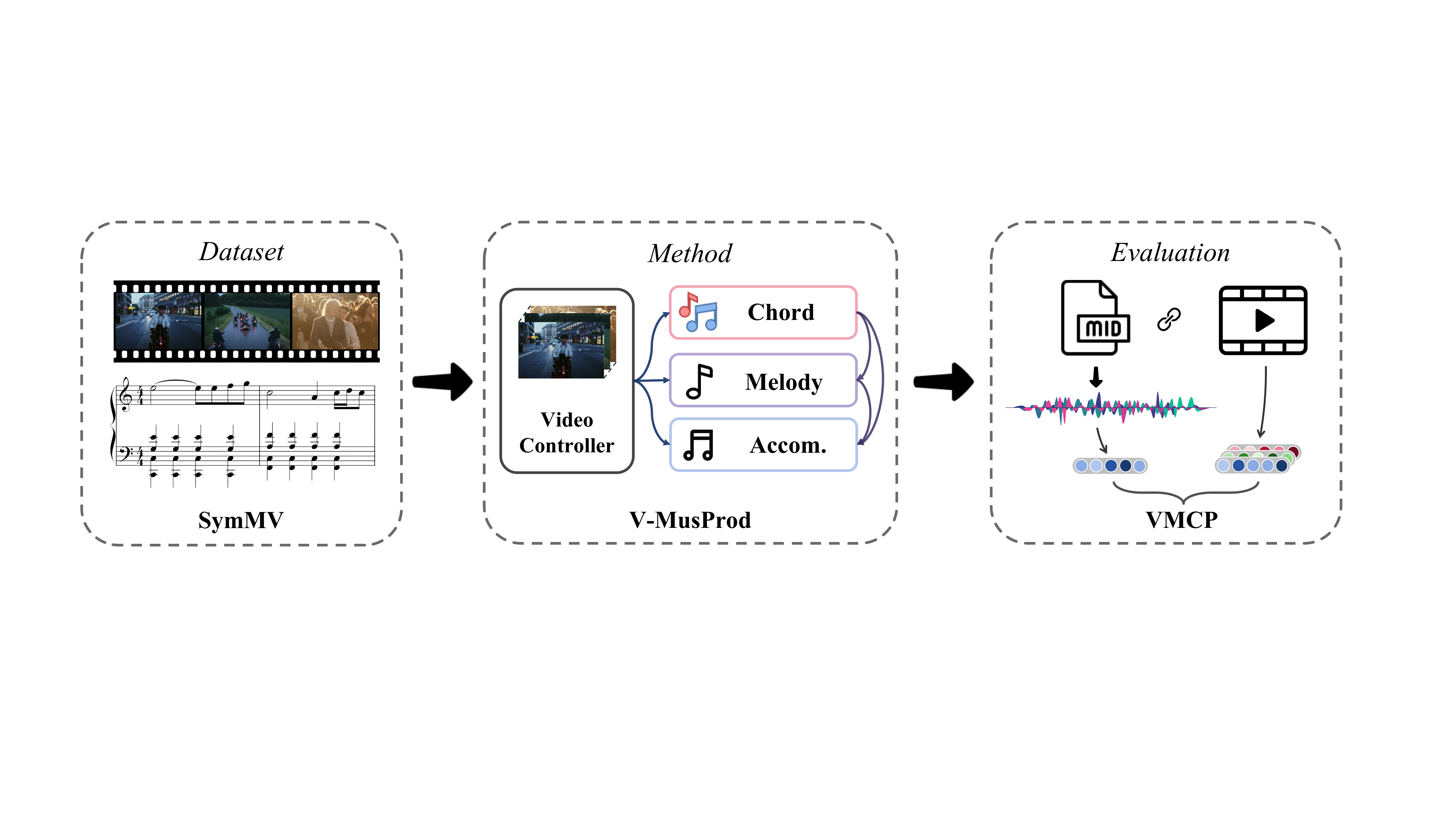}
     \vspace{-2mm}
    \caption{\textbf{Overview of our framework.} We solve the task of video background music generation from three perspectives. \textbf{Left:} We present the first video and symbolic music dataset with detailed annotations. \textbf{Middle:} We decouple music generation into three progressive stages: chord, melody, and accompaniment (accom.), and extract various video features to guide different generation stages. \textbf{Right}: We propose a novel evaluation metric, named Video-Music CLIP Precision (VMCP), to measure the correspondence between generated music and input video.}
    \label{fig:overview}
    \vspace{-5mm}
\end{figure*}

Recently, tremendous progress has been made with text-to-image systems~\cite{dalle, imagen}, revealing the powerful generative capacity of state-of-the-art models. Though available to the same family of generative models, the area of video-to-music generation is still in the preliminary stage. We attribute this to the following three key challenges from the aspects of dataset, method, and evaluation, respectively: (1) Large-scale datasets of high-quality music with paired videos are absent and cumbersome to collect. (2) Video-music correspondence is complex and tricky to integrate effectively into current generative models. (3) There is a lack of objective metrics to evaluate the correspondence between video and music. We give detailed explanations and provide the corresponding solutions for each challenge in the rest of our paper. Our framework is illustrated in Fig.~\ref{fig:overview}.

Existing video-music datasets~\cite{vm-net, aist++, d2m-gan} are either limited in size or weak in video-music correspondence due to noisy data pairs. More importantly, these datasets only include music in audio format, which is complex, computationally expensive, and difficult to impose control signals from videos. On the contrary, symbolic music, representing music in discrete sequence~\cite{remi, cp}, contains rich semantic information and helps to explore video-music relationships.

To fill this gap, we introduce a novel video and symbolic music dataset named Symbolic Music Videos (SymMV). It contains piano covers of popular music with their official music videos carefully collected from the Internet. Overall, it contains 1140 video-music pairs of more than 10 genres with a total length of 76.5 hours. As an advantage of symbolic music, we provide chord, melody, accompaniment, and other metadata for each music. The detailed annotations allow model to decouple the music generation process into stages and better control the generated music. Note that music in SymMV is of high quality and can also be directly used for unconditional music generation without video modality.

We then explore the method for video background music generation. It is challenging since the correspondence between music and video is not a deterministic one-to-one mapping but a more complex one related to aesthetic style. Models are required to create music that is not only coherent and melodious but also harmonic with the given video in terms of both rhythm and style. Some initial attempts~\cite{how_does_it_sound, d2m-gan} solve the motion-to-music generation via rhythmic control by motion features in videos. They suffer from serious limitations on applied video types,~\emph{i.e.}, requiring additional key points annotation. Prior work CMT~\cite{cmt} designs rule-based rhythmic relationships to generate video background music due to lack of paired data. It ignores the semantic-level correspondence, which sometimes results in conflicting styles.

To build a benchmark model, we propose a \underline{V}ideo \underline{Mus}ic generation framework with \underline{Pro}gressive \underline{d}ecoupling control (V-MusProd). 
V-MusProd decouples music generation into three progressive transformer stages: chord, melody, and accompaniment. 
It first predicts a chord sequence, then generates melody conditioned on chords and finally generates accompaniment conditioned on chords and melody. 
We extract semantic and color features to control the Chord Transformer since these features reflect the theme and emotion of a video. Motion features are extracted as rhythmic control for Melody and Accompaniment Transformers since their outputs require rhythmic alignment with the input video.  
V-MusProd can generate melodious music corresponding to the input video based on the two controlling processes. 

Lastly, another bottleneck of video background music generation lies in the inadequate objective metrics. Previous methods~\cite{cmt, d2m-gan, how_does_it_sound} use metrics for unconditional music generation or subjective evaluation to examine the video-music correspondence. Common metrics for unconditional music generation can only assess the quality and diversity of the generated samples but ignores the alignment between paired music and videos. Besides, conducting human listening tests for subjective evaluation is cumbersome and sometimes biased. 
Recent text-to-image generation models~\cite{glide, imagen} leverage the powerful pretrained vision-language CLIP~\cite{clip} model to compute the similarity between input prompts and generated images, opening up new opportunities for evaluating sample correspondence. Therefore, we propose a new evaluation metric, named Video-Music CLIP Precision (VMCP), which extends the vision-language CLIP model to video and music domain to measure the video-music correspondence. With VMCP and subjective evaluation, V-MusProd demonstrates better results than CMT on SymMV.

Our contributions are summarized as follows: (1) We present SymMV, the first video and symbolic music dataset with detailed annotations tailored for video background music generation.
(2) We propose V-MusProd, a benchmark framework that utilizes music priors of chords, melody, and accompaniment along with video-music relations of semantic, color, and motion features. 
(3) We propose an objective metric VMCP based on the video-music CLIP model. Both objective metrics and subjective evaluation demonstrate that V-MusProd surpasses the state-of-the-art model in video-music correspondence and music quality.

\vspace{-1mm}
\section{Related Work}
\label{sec:related}
\vspace{-1mm}

\noindent\textbf{Video-music Datasets.}
Multi-modal datasets,~\emph{e.g.}, image-text~\cite{laion, conceptual, wit}, video-text~\cite{howto100m, youcook2}, video-audio~\cite{vggsound, audioset, acav100m, aist++} greatly push the development of multi-modal learning tasks. However, there lacks datasets for video-conditional music generation. 
HIMV~\cite{vm-net} is a large-scale dataset with 200K video-music pairs from YouTube-8M~\cite{youtube-8m}. Despite its size, it suffers from weak correspondence and poor quality of music and videos,~\emph{e.g.}, static videos, live performances, or amateur montages, since it is only designed for the task of retrieval. 
Recently, AIST++~\cite{aist++} and TikTok dataset~\cite{d2m-gan} are introduced for music-dance learning, which contains videos, motion sequences, and paired music. However, their music parts are limited in size,~\emph{i.e.}, less than 5 hours, and are only provided in audio format.
Therefore, it is important to build a video-music dataset tailored for music generation. 
Referring to symbolic music generation literature\cite{pop909, music_transformer, cp}, we build SymMV, which includes 1140 piano music in MIDI format with paired music videos and rich musical annotations. 

\noindent\textbf{Video-Conditional Music Generation.}
Most music generation methods fall into the unconditional setting~\cite{jukebox, music_transformer, remi, cp}. Previous video-conditional music generation works mainly focus on reconstructing music from silent instrument performance videos~\cite{multi_instrument_net, foley_music, audeo, sight2sound}. Recent methods ~\cite{how_does_it_sound, d2m-gan, cdcd} are proposed to generate music for dancing or human activity videos based on rhythmic relations. However, their methods require extra motion annotations as inputs and thus do not work for general videos with various contents. CMT~\cite{cmt} generates background music for general videos with video rhythmic features, but its video-music relationships are purely rule-based, and it lacks semantic relationships between videos and music. 

\noindent\textbf{Objective Music Metrics.}
Previous objective music metrics~\cite{musegan, muspy, ji2020comprehensive} mainly focus on statistics of music features, where the performance of generated music is decided by their closeness to the training data. 
Text-to-image generative models use R-precision~\cite{attngan} to evaluate whether the generated image can be used to retrieve its input text description. Recent works~\cite{clip_r_precision, tise} improve the robustness of R-precision using the powerful CLIP model~\cite{clip}. Cosine similarity computed by CLIP, named CLIPScore~\cite{clipscore}, can also be directly used as an objective metric.

\vspace{-1mm}
\section{Dataset}
\label{sec:dataset}
\vspace{-1mm}

\begin{table*}
\small
 \resizebox{1.0\textwidth}{!}{
    \centering
    \begin{tabular}{lcccccccccc}
    \bottomrule
    \multirow{1}{*}{Dataset} & \multirow{1}{*}{Video} & \multirow{1}{*}{Audio} & \multirow{1}{*}{MIDI} & \multirow{1}{*}{Genre} & \multirow{1}{*}{Chord} & \multirow{1}{*}{Melody} & \multirow{1}{*}{Tonality} & \multirow{1}{*}{Video Content} & \multirow{1}{*}{Size} & \multirow{1}{*}{Length (Hours)}\\ \midrule
    MAESTRO~\cite{maestro}  & \XSolidBrush & \Checkmark & \Checkmark & \XSolidBrush & \XSolidBrush & \XSolidBrush & \XSolidBrush & - & 1,276 & 198.7 \\
    POP909\cite{pop909} & \XSolidBrush & \XSolidBrush & \Checkmark & \Checkmark & \Checkmark & \Checkmark & \Checkmark & - & 909 & 70.0 \\ \midrule
    HIMV-200K\cite{vm-net} & \Checkmark & \Checkmark & \XSolidBrush & \XSolidBrush & \XSolidBrush & \XSolidBrush & \XSolidBrush & Music Video & 200,500 & - \\
    TikTok\cite{d2m-gan} & \Checkmark & \Checkmark & \XSolidBrush & \XSolidBrush & \XSolidBrush & \XSolidBrush & \XSolidBrush & Dance Video & 445 & 1.5 \\ 
    AIST++\cite{aist++} & \Checkmark & \Checkmark & \XSolidBrush & \Checkmark & \XSolidBrush & \XSolidBrush & \XSolidBrush & Dance Video & 1,408 & 5.2  \\
    URMP\cite{urmp} & \Checkmark & \Checkmark & \Checkmark & \XSolidBrush & \XSolidBrush & \XSolidBrush & \XSolidBrush & Music Performance & 44 & 33.5 \\
    \midrule
    SymMV (Ours) & \Checkmark & \Checkmark & \Checkmark & \Checkmark & \Checkmark & \Checkmark & \Checkmark & Music Video & 1,140 & 76.5 \\ \bottomrule
    \end{tabular}
    }
\vspace{-1mm}
\caption{\textbf{Comparison between different music datasets.} The proposed SymMV is the first dataset includes video and symbolic music pairs for video background music generation. Our dataset also provides various musical annotations and metadata such as genre, chord, and melody. We include popular symbolic music datasets in the first two rows for reference.}
\label{tab:dataset}
\vspace{-3mm}
\end{table*}

\begin{figure}[t]
    \centering
    \includegraphics[width=\linewidth]{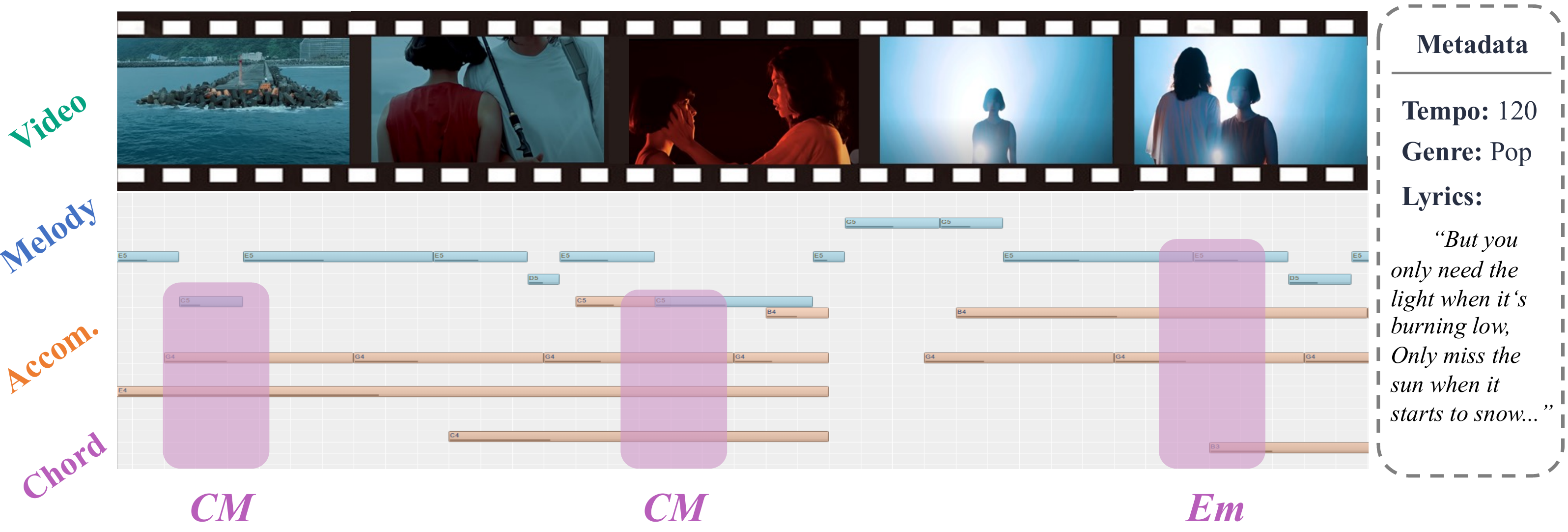}
    \vspace{-3mm}
    \caption{\textbf{Illustration of sample in SymMV.} Sample in our SymMV dataset includes paired music and video, music feature annotations, and related metadata.}
    \label{fig:dataset}
    \vspace{-3mm}
\end{figure}

We collect the first video-music dataset that matches music videos and their piano version music in symbolic form. The collected SymMV dataset contains 1140 pop piano music in both MIDI and audio format with the corresponding official music video with a total duration of 76.5 hours. We split SymMV into the training set (1000 pairs), validation set (70 pairs), and test set (70 pairs). Our dataset also includes various annotated metadata, such as chord progression, tonality, and rhythm. Fig.~\ref{fig:dataset} shows an example of our dataset.

Tab.~\ref{tab:dataset} provides comparisons with existing video-music datasets. 

\vspace{-1mm}\subsection{Data Collection}
\label{sec:data-collection}
\vspace{-1mm}
Mass amounts of video-music pairs are available on the Internet. In particular, music videos have strong video-music correspondence in artistic styles and rhythms. They also have plenty of scenes, movements, and camera angles, thus suitable for learning intrinsic video-music relations. Therefore, we aim to construct a music video and paired symbolic music dataset for video background music generation.

Although finding music videos or their piano covers separately can be easy, it is hard to collect corresponding pairs. We solve this challenge by first collecting piano covers with fair to high audio and musical quality from professional piano tutorial YouTube channels. After downloading both audio and their metadata, we parse the metadata and use the parsed song title and singer as keywords to search for the corresponding official music video. Once video and audio pairs are collected, we transcribe the audio files into MIDI format using a state-of-the-art automatic piano transcription model~\cite{onsets_and_frames}. To ensure dataset quality, we invite three professional musicians to manually check the collected dataset. They filter out low-quality pairs, such as static or lyric videos and fragmented or undesirable MIDIs, and find the resulting accuracy in note pitch and duration satisfactory.

\vspace{-1mm}\subsection{Data Annotations}
\label{sec:data-annotation}
\vspace{-1mm}

\noindent\textbf{Melody and Accompaniment.}
Common pop music is well structured and can be decoupled into melody and accompaniment.
\emph{Melody}, a combination of pitch and rhythm, constitutes the most memorable aspect of a song. It is easier for people to perceive melody than other music parts. Hence, melody plays an essential role in a music piece.
\emph{Accompaniment} is correlated with the chord sequence and melody, serving as the background sound effect to harmonize the foreground melody~\cite{tonal_harmony}, and can bring different auditory sensations. 
Given music note sequences, we first quantize their duration to the 16th note and implement the Skyline algorithm~\cite{skyline} to separate the melody and accompaniment.

\noindent\textbf{Chord Progression.}
\emph{Chords}, several notes in a certain vertical configuration that sounds harmonic, run through the whole music piece and play an important role in setting the base tone of music. Moreover, chords convey strong emotions with several unique features like color and tension,~\emph{e.g.}, major chords bring a feeling of brightness, while minor chords sound relatively dim~\cite{schoenberg1983theory}. We provide chord progression to further control the music generation process. We adopt an open-source rule-based algorithm~\footnote{https://github.com/joshuachang2311/chorder/} to extract chords from MIDI files. We observe a long-tail distribution of chord progressions, where more than half of the chords occur less than 10 times. To mitigate this problem, we narrow down the chord templates to 12 root notes and 10 qualities, which covers mostly used types in pop songs.

\noindent\textbf{Tonality.}
\emph{Tonality} is the general term for \emph{tonic} and \emph{mode} of a key. It reflects the hierarchy of stability, attractions, and directionality in music work. The tonic chord is considered to be the most stable chord in tonality, and it determines the name of the key. Mode represents a type of musical scale centered on the tonic, which can be mainly divided into two types: major and minor modes. In our dataset, we provide the tonality annotation using Krumhansl-Schmuckler algorithm~\cite{krumhansl2001cognitive} to predict tonality from MIDI files and represent music keys using 12 tonic and 2 mode types.

\noindent\textbf{Rhythm.}
We estimate the beat and downbeat positions from audio using the RNN-based model~\cite{madmom}, which corresponds to the fine-grained rhythm. Then, we calculate the tempo from beat and downbeat positions to represent the global rhythm.

\noindent\textbf{Metadata.}
We also provide additional metadata of our music dataset, such as genre and lyrics. We use ShazamIO\footnote{https://github.com/dotX12/ShazamIO} to search for lyrics and genres of music in SymMV. These metadata are helpful for data analysis and may benefit future applications,~\emph{e.g.}, text-to-music generation~\cite{musiclm, noise2music}.

\begin{figure}[t]
    \centering
    \includegraphics[width=1.0\linewidth]{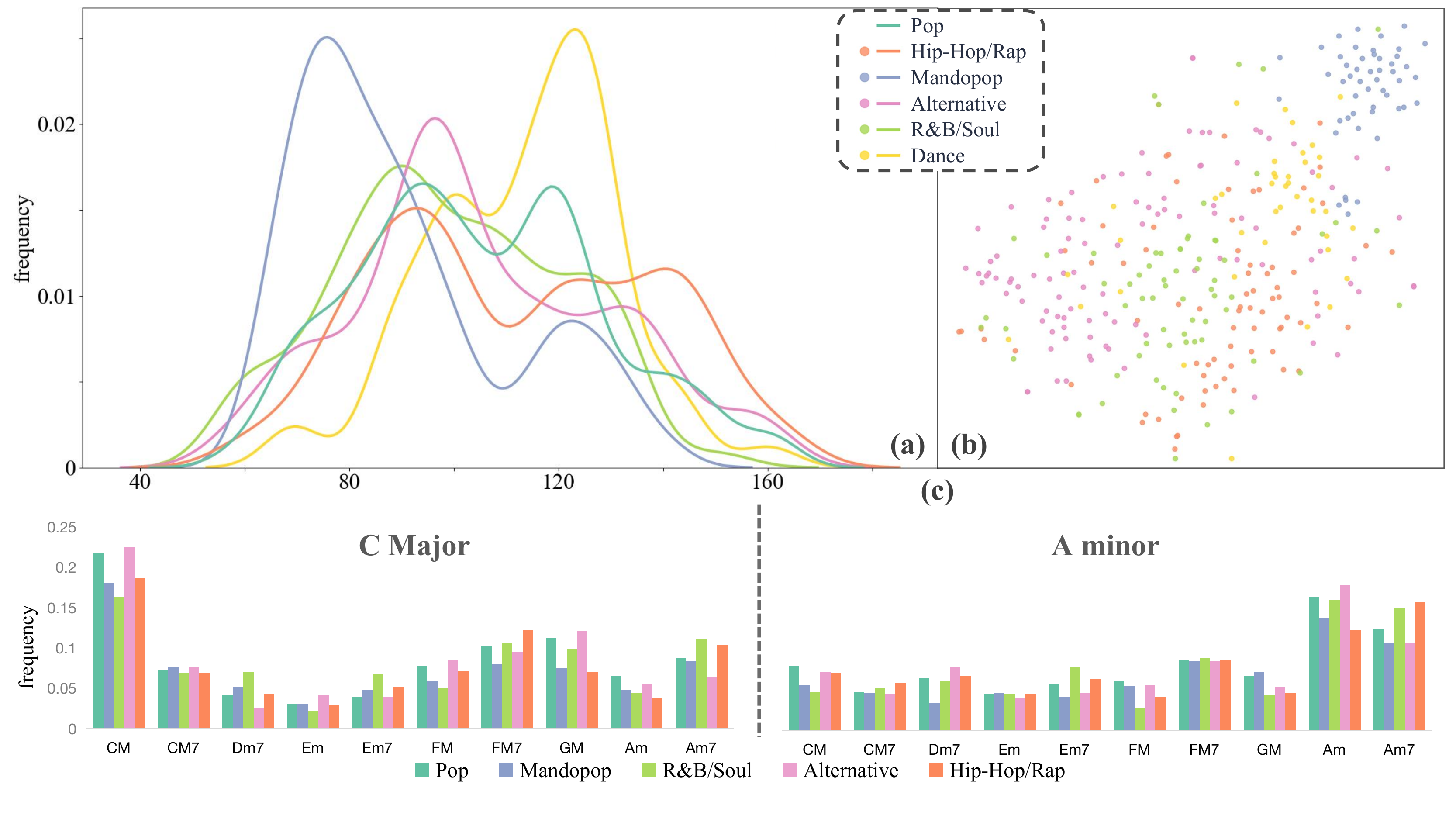}
    \vspace{-3mm}
    \caption{\textbf{Visualization of data statistics.} (a) Probability density function of beats per minute in different genres. (b) T-SNE visualization of visual features and genres. (c) Chord distribution in different genres. We show the high correlation between not only music and genre but also video and genre.}
    \vspace{-3mm}
    \label{fig:Anylasis}
\end{figure}

\subsection{Data Analysis}
\label{sec:data-analysis}
\vspace{-1mm}
To ensure the quality of our dataset and determine video-music relationships
, we provide a detailed analysis of different music and video features. Genre, an attribute shared by both modalities, is convenient for us to analyze with visualization tools. Since our dataset contains video-music pairs with more than 10 genres, we use genre as a bridge between video and music to explore their distinct features. We transpose all major music to C major (C) and minor music to A minor (a) to remove the influence of tonality on our analysis.

\noindent\textbf{Chord and Genre.}
We count the frequency of chords in different genres of music. To ensure statistical significance, we choose the five most frequent genres and ten chord types with high frequency and high variance. As Fig.~\ref{fig:Anylasis} (c) shows, the final chord distribution meets our expectations. In \emph{pop} music, the most steady chords, \emph{e.g.}, CM, FM, and Am, occupy the largest proportion, in line with the stability of pop music. In contrast, some rarely seen chords have a high frequency in \emph{R\&B/Soul}, such as Dm7 and Am7, in order to create a richer and more diverse harmonic palette. \emph{Alternative}, as a type of \emph{Rock}, tends to favor simpler chords, so the occurrences of seventh chords are less frequent. As for \emph{Hip-Hop/Rap}, the singing part is less melodic, and therefore, there are generally more seventh chords to provide accompaniments with more space for expression and to fill the harmony space. We provide more analysis in the Appendix.

\noindent\textbf{Rhythm and Genre.}
We use kernel density estimate (KDE) to visualize the distribution of beats per minute (BPM) of music in various genres. As shown in Fig.~\ref{fig:Anylasis} (a), ~\emph{Dance} and~\emph{Hip-Hop/Rap} tend to have higher BPM. Curves in other genres display two distinct peaks, representing the BPM of slow-paced and fast-paced songs, respectively. Notably, there is no obvious peak corresponding to fast songs in~\emph{R\&B/Soul}.

\noindent\textbf{Visual Features and Genre.}
As for visual features, we first extract 512-dimensional CLIP features from video frames at an FPS of 6. Then we compute the average of these features at time dimension to generate the visual feature of the whole video. We use t-SNE~\cite{tsne} to project visual features into a 2-dimensional space. We ignore \emph{Pop} due to its complexity and select CLIP features of the other five genres to conduct the cluster analysis. Visual features are generally clustered by genre in Fig.~\ref{fig:Anylasis} (b), demonstrating high correlations.

\vspace{-1mm}
\section{Method}
\label{sec:method}
\vspace{-1mm}

\begin{figure*}[t]
    \vspace{-2mm}
    \centering
    \includegraphics[width=0.9\textwidth]{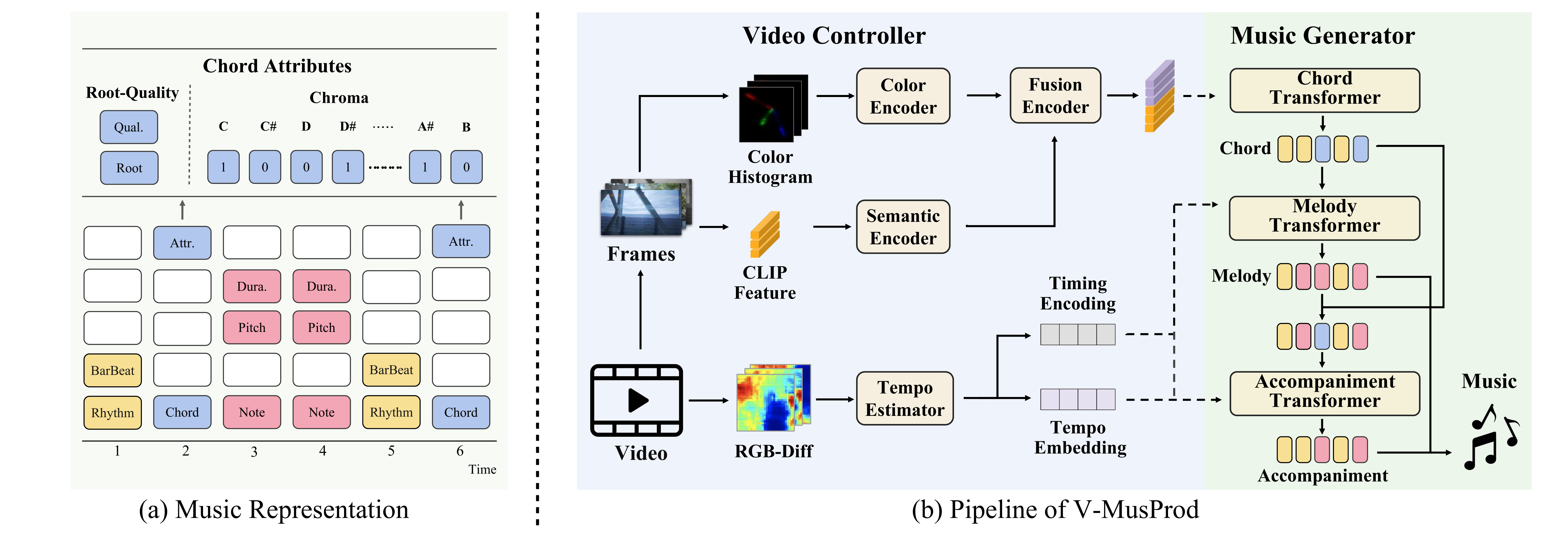}
    \vspace{-3mm}
    \caption{\textbf{Illustration of our method V-MusProd.}
    \textbf{Left}: We group multiple music attributes into one event-based token (each column). Different colors indicate different types of tokens. \textbf{Right}: We extract semantic, color, and motion features to guide the music generation process. The three types of features serve as inputs for different stages of the decoupled music generation model, which contains Chord, Melody, and Accompaniment Transformers.}
    \vspace{-3mm}
    \label{fig:video}
\end{figure*}

We propose a novel music generation framework named V-MusProd, to tackle the challenging video background music generation task. Our framework is shown in Fig.~\ref{fig:video}, which consists of a video controller and a music generator. The video controller extracts visual and rhythmic features and fuses them as the contextual input of the music generator. The music generator decouples the music generation process into three progressive stages that are independently trained: Chord, Melody, and Accompaniment. At inference time, melody and accompaniment tracks are merged together to form a complete music piece. The progressive generation pipeline allows for the use of decoupling control on different generation stages, which improves the correspondence between videos and music.
We elaborate on each component in the following subsections.

\vspace{-1mm}\subsection{Video Controller}
\label{sec:video-controller}
\vspace{-1mm}
Directly using raw video frames as conditional input is difficult for model to learn the correspondence between two different modalities. Thus, it is important to design and extract meaningful features from video as intermediate representations to simplify the learning process. 
Considering the style and rhythm relationship between music and video, we extract semantic, color, and motion features separately to guide the music generation model.

\noindent\textbf{Semantic Features.} We use pretrained CLIP2Video~\cite{clip2video} as the extractor to encode raw video frames into semantic feature tokens without finetuning. The CLIP2Video model builds upon the CLIP encoder~\cite{clip}, which is pretrained on billions of image-text pairs, and further uses a temporal difference block to learn the temporal context across frames. The extracted features are supposed to contain representations of different video semantics,~\emph{e.g.}, scenery, sports, and crowds, which are closely related to the content of music. 

\noindent\textbf{Color Features.} Color in videos can reflect the underlying emotions in a given scene, corresponding with the mood of paired music. We employ color features as one of the control signals for chord generation. Specifically, we extract the color histogram of each video frame, \emph{i.e.}, a 2D feature map proposed in~\cite{histogan}, to represent the color distribution in a non-linear manifold. The color histogram projects an image's color into a log-chroma space, which is more robust and invariant to illumination changes.

Semantic and color features are fed into separate transformer encoders and then concatenated together at the length dimension. We add a learnable embedding to mark whether each token is from color feature or semantic feature, and feed the sequence into a transformer encoder for inter-modality and temporal fusion. The fused output serves as keys and values of cross attention in Chord Transformer.

\noindent\textbf{Motion Features.} We compute RGB difference as motion features to determine the music tempo and calculate Tempo Embedding and Timing Encoding. 
We extract the RGB difference with intervals of $5$ frames ($0.2$ seconds) and map the mean RGB difference of a video to the music tempo. We use a linear projection from the minimum and maximum RGB difference to the tempo range of $[90, 130]$.
The estimated tempo is used as the Tempo Embedding in the music generator. We also add Timing Encoding~\cite{cmt} in Melody and Accompaniment Transformers to synchronize the video timing and music timing, which reminds the model of the current token's position in the whole sequence. 

\vspace{-1mm}\subsection{Music Generator}
\label{sec:video-generator}
\vspace{-2mm}
The music generator $G$, consisting of a Chord Transformer $G_c$, a Melody Transformer $G_m$, and an Accompaniment Transformer $G_a$, is designed to generate symbolic music conditioned on the extracted video feature. The workflow can be written as follows: 
\begin{equation*}
\small
\begin{aligned}
x_m =& G_m(G_c(y_s), y_r), \\
x_a =& G_a(G_c(y_s), x_m, y_r), \\
x =& x_m \oplus x_a,
\end{aligned}
\end{equation*}
where style feature $y_s$ and rhythm feature $y_r$ is produced by video controller $C$, the final music piece $x$ is composed of the melody $x_m$ and the accompaniment $x_a$, $\oplus$ represents merging the two parts into a single track. 

\noindent\textbf{Music Representation.}
Symbolic music comprises a set of music attributes. To encode the dependencies among different attributes, we design an event-based music representation inspired by~\cite{cp, popmag}. We define three types of tokens, namely {Note}, {Rhythm}, and {Chord}, as the red, yellow, and blue columns in Fig.~\ref{fig:video} (a), respectively. Each token is a stack of attributes. In particular, the Rhythm token comprises the BarBeat attribute, indicating the beginning of each bar or beat; Note token contains the Pitch and Duration attributes; Chord token contains the Root and Quality attributes, \emph{i.e.}, the root note and the quality of chords. Chord can also be represented as chromagrams, a $12$D binary vector where each dimension indicates whether a pitch class is activated. An additional Type attribute is applied for all tokens to mark their types. In our implementation, the Chord Transformer only models the Rhythm and the Chord tokens, while the Melody and Accompaniment Transformers model all three token types. To align the generated music with input video with rhythmic information, we add Bar Embedding and Beat Embedding for the absolute bar and beat position of current token, and Tempo Embedding for the music tempo. 

\noindent\textbf{Chord Transformer.}
We adopt a transformer decoder architecture for Chord Transformer to learn the long-term dependency of input video feature sequences. The event-based token sequence is added with positional encoding and fed into the Chord Transformer as a query. Meanwhile, the style features from video controller are fed as keys and values.

\noindent\textbf{Melody Transformer.}
We employ an encoder-decoder transformer architecture for melody generation. The encoder receives a chord sequence as input, and then the decoder generates a note sequence as the output melody. 
Considering the relatively short-range dependency between melody and chords, we adopt a bar-level cross-attention mask so that each decoder token can only attend to the contextual encoder output within the previous current or next bar.

\noindent\textbf{Accompaniment Transformer}
Similarly, we also adopt an encoder-decoder transformer to generate the accompaniment sequence. Since accompaniment closely correlates with chords and melody, we merge the generated chord sequence with the melody and then pass the merged sequence to Accompaniment Transformer as conditional input. We also apply the same bar-level cross-attention mask as in Melody Transformer. Eventually, the generated accompaniment is directly merged with the melody to form the final music.

\vspace{-1mm}
\subsection{Implementation Details}
\vspace{-1mm}
\label{sec:implementation}
We train three stages separately and connect them to form a complete pipeline during inference. We construct transformers~\cite{transformer} based on linear transformer~\cite{linear_transformer} to reduce time consumption. All three stages are trained with cross-entropy loss and teacher-forcing strategy. 
During inference, we use a stochastic temperature-controlled sampling~\cite{sampling} to increase the diversity of generated samples. We train Chord, Melody, and Accompaniment Transformers for 200, 200, and 400 epochs, respectively, on one V100 GPU.
We use fluidsynth\footnote{https://github.com/FluidSynth/fluidsynth} to synthesize our MIDI into audio.
More implementation details are in the Appendix.

\vspace{-1mm}
\section{Evaluation Metric}
\label{sec:metrics}
\vspace{-1mm}
In this chapter, we first extend the vision-language CLIP~\cite{clip} to the video-music domain and propose a new evaluation metric named Video-Music CLIP Precision (VMCP) to measure the video-music correspondence. 

\vspace{-1mm}\subsection{Video-Music CLIP}
\label{sec:clip}
\vspace{-1mm}
To build the video-music CLIP model, we adopt the design choice in~\cite{musicforvideo}, the state-of-the-art video-music retrieval model. Specifically, we first split the input music and video into fixed-length segments and use CLIP~\cite{clip} and music tagging model~\cite{music_tagging} to extract visual and audio features separately. Given the extracted features, we adopt a transformer encoder as the video encoder and music encoder to explore contextual relations and learn a joint multi-modal embedding space. The model is trained with the InfoNCE contrastive loss~\cite{infonce} to map positive video-music pairs closer while pushing negative pairs further in the CLIP-based joint embedding space. Loss of videos ${v}$ to music pieces ${m}$ is defined as:
\begin{equation*}
\small
    \mathcal{L}^{v\rightarrow{}m} = - \sum_i^N\sum_l^L\left[\log\frac{\exp\left[s(v_{i,l},m_{i,l})/\tau\right]}{{\sum_{j}^N\sum_k^L\exp{\left[s(v_{i,l},m_{j,k})/\tau\right]}}}\right],
    \label{eq:infonce}
\end{equation*}
where $N$ denotes number of video-music pieces, $L$ denotes number of segments, $s(\cdot)$ denotes cosine similarity,
and $\tau$ is a learnable temperature parameter. The music-to-video loss $\mathcal{L}^{m\rightarrow{}v}$ is defined symmetrically.

To train this model, we need to collect plenty of video-music pairs and ensure that the training dataset should roughly cover the distribution of our dataset. Therefore, we download video clips from YouTube8M dataset~\cite{youtube-8m} annotated as ``music video"
and obtain $\sim$20k video-music pairs. After training on the YouTube music video dataset, we finetune the model with a small learning rate on audio converted from SymMV to improve its retrieval performance further.

\vspace{-1mm}\subsection{Video-Music CLIP Precision (VMCP)} 
\label{sec:clip-precision}
\vspace{-1mm}
Equipped with the pretrained video-music CLIP model, we design a retrieval-based metric similar to~\cite{attngan}. Given a generated music piece in MIDI format, we first synthesize it into audio and calculate the top-K retrieval accuracy from a pool of N candidate videos using the CLIP model. Specifically, we rank the cosine similarity between the generated sample $\hat m$ and its condition video $v$ and $M-1$ random sampled videos $v_i$. We consider a successful retrieval if the ground truth video is ranked in the top-$K$ place. We test the model using all generated samples and compute the success retrieval rate as the final precision score. We set $M=70, K=5, 10, 20$. We also calculate the average rank of the ground truth video, where a lower rank implies better correspondence. Overall, the proposed metric is able to measure how well the generated music aligns with the input video. We validate that it shows a high correlation with human judgments in the experiments.

\begin{table*}
\small
    \begin{center}
    \begin{tabular}{lccccccccc}
        
        \toprule
        \multirow{2}{*}{Methods} & \multicolumn{4}{c}{Video-Music Correspondence} & \multicolumn{5}{c}{Music\ Quality} \\

        & P@5 & P@10 & P@20 & AR & SC & PE & PCE & EBR & IOI \\ \midrule
        
        Real (SymMV) & - & - & - & - & 0.986 & 4.197 & 2.633 & 0.023 & 0.184 \\ \midrule

        CMT~\cite{cmt} & 8.9 & 17.7 & 31.0 & 33.4 & \underline{0.990} & 3.920 & 2.444 & 0.074 & 0.246 \\ \midrule

        w/o semantic & 11.6 & 23.9 & 42.0 & 26.1 & 0.955 & 2.892 & 2.310 & \textbf{0.019} & 0.358 \\
        
        w/o color & \underline{15.6} & \textbf{26.6} & \textbf{44.8} & \textbf{25.1} & 0.956 & 2.732 & 2.200 & \underline{0.011} & 0.330 \\
        
        w/o motion & 12.2 & 22.2 & 37.9 & 26.3 & 0.975 & 3.010 & 2.283 & 0.004 & 0.261 \\
        
        Video2music & 10.8 & 19.7 & 33.3 & 30.0 & 0.981 & \textbf{3.990} & \textbf{2.639} & 0.010 & \underline{0.229} \\
        
        Video2chord2music & 13.7 & 23.1 & \underline{43.6} & 26.0 & 0.996 & 2.497 & 2.036 & 0.081 & 0.985 \\
        
        V-MusProd & \textbf{15.7} & \underline{24.6} & \textbf{44.8} & \underline{25.4} & \textbf{0.983} & \underline{3.940} & \underline{2.607} & 0.004 & \textbf{0.174} \\
        \bottomrule

    \end{tabular}
    \end{center}
\vspace{-3mm}
\caption{\textbf{Objective evaluation on SymMV test set.} We evaluate video-music correspondence and music quality with VMCP and music quality metrics. P indicates Precision, where higher is better. AR indicates average rank, where lower is better. For music quality metrics, \textbf{closer} to Real is better.
}
\label{tab:cmt}
\vspace{-4mm}
\end{table*}

\vspace{-1.5mm}
\section{Experiments}
\label{sec:exp}
\vspace{-1.5mm}

We conduct comprehensive experiments 
on our V-MusProd model. In Sec.~\ref{exp:compared}, we introduce the compared method CMT~\cite{cmt}. 
In Sec.~\ref{exp:objective}, we evaluate video-music correspondence with VMCP and music quality with objective metrics. 
In Sec.~\ref{exp:subjective}, we conduct a thorough subjective evaluation for video-music correspondence and music quality by user study.
In Sec.~\ref{exp:ablation}, we ablate our design choices and highlight the importance of different features used in video controller to validate the effectiveness of proposed method.
In Sec.~\ref{exp:uncond}, we train V-MusProd in unconditional setting and evaluate its music quality against previous symbolic music generation methods.

\vspace{-1mm}
\subsection{Compared Method}
\label{exp:compared}
\vspace{-1mm}
We compare V-MusProd with the state-of-the-art video background music generation method CMT~\cite{cmt}, the first and only method to generate full-length background music for general videos. CMT uses purely rule-based video-music rhythmic relationships without paired video-music data. Other video-conditional music generation methods mostly focus on specific video types (\emph{e.g.} dance videos) and require extra annotations (\emph{e.g.} keypoints~\cite{d2m-gan, how_does_it_sound}), which are unavailable in the general setting. We train CMT on SymMV to provide a benchmark.

\vspace{-1mm}
\subsection{Objective Evaluation}
\label{exp:objective}
\vspace{-1mm}

\noindent\textbf{Metrics.} 
For video-music correspondence, we use VMCP to evaluate objectively, where higher precision and lower average rank are better. For music quality, we select music objective metrics from~\cite{muspy, mgeval}, including
Scale Consistency (SC), Pitch Entropy (PE), Pitch Class Entropy (PCE), Empty Beat Rate (EBR), and average Inter-Onset Interval (IOI), which evaluate music by pitches and rhythm. Note that these music quality metrics are not indicated by how high or low they are but instead by their \emph{closeness} to the real data. We use SymMV test set for evaluation.

\noindent\textbf{Results.} 
As shown in Tab.~\ref{tab:cmt}, V-MusProd surpasses CMT on VMCP and music quality metrics. This proves our method achieves better video-music correspondence and music quality than the state-of-the-art method.

\vspace{-1mm}
\subsection{Subjective Evaluation}
\label{exp:subjective}
\vspace{-1mm}
Subjective evaluation is widely adopted in previous works~\cite{cmt, cp, hat, music_transformer}. We conduct a user study by sending out questionnaires. We invite 55 participants, including 10 \emph{experts} with expert knowledge in music composition and 45 \emph{non-experts}.
We provide several videos from different categories like scenery, city scenes, and movies. Each video has two pieces of background music generated by V-MusProd and CMT, presented randomly for blindness. The questionnaire takes about 20 minutes to complete.

\noindent\textbf{Metrics}. Participants are required to compare two background music pieces from several aspects: (1)~Music Melody: melodiousness and richness of music theme; (2)~Music Rhythm: structure consistency of rhythm; (3)~Video Content: correspondence between video content and music; (4)~Video Rhythm: correspondence between video motion and music rhythm; (5)~Overall Ranking: overall preference of the two samples. Besides, we ask the expert group to evaluate two additional metrics related to music theory: (6)~Chord Quality: the quality of chord progression in the music; (7)~Accompaniment Quality: the richness of music accompaniment.

\noindent\textbf{Results}. We provide results of preference rate, \emph{i.e.} the percentage of users who consider our music better than CMT, in Tab.~\ref{tab:subjective}. Results show that V-MusProd outperforms CMT ($>50\%$) in nearly all metrics and user groups, demonstrating better music quality and correspondence with videos. In particular, our model outperforms CMT in both Music Melody and Accompaniment Quality by a large margin, indicating our decoupling generation of melody and accompaniment significantly improves their qualities. The subjective evaluation results show high correlations with VMCP, which further verifies the effectiveness of our proposed metric.

\begin{table}
\small
    \begin{center}
    \begin{tabular}{cccc}
        \toprule
        Metrics & Expert & Non-expert \\
        \midrule
        Music Melody & 77\% & 82\% \\
        Music Rhythm & 63\% & 53\% \\
        Video Content & 63\% & 63\% \\
        Video Rhythm & 60\% & 57\% \\
        Chord Quality & 63\% & - \\
        Accom. Quality & 83\% & - \\
        Overall Ranking & 73\% & 67\% \\
        \bottomrule
    \end{tabular}
    \end{center}
\vspace{-4mm}
\caption{\textbf{Subjective evaluation for V-MusProd against CMT~\cite{cmt}.} We show preference rates in music quality metrics, video-music correspondence metrics, and expertise metrics.}
\label{tab:subjective}
\vspace{-4mm}
\end{table}

\vspace{-1mm}
\subsection{Ablation Study}
\label{exp:ablation}
\vspace{-1mm}

\noindent\textbf{Ablation on Video Controller.} 
We ablate the three video control features and evaluate the video-music correspondence with VMCP: (1) w/o semantic: no semantic feature input for video controller; (2) w/o color: no color feature input for video controller; (3) w/o motion: fix tempo and do not add timing encoding.
As shown in Tab.~\ref{tab:cmt}, semantic and motion features are significant for correspondence. We observe that w/o color has similar correspondence with the full model despite lower music quality. We attribute this to the fact that music tonality is connected with the color of videos. The original keys have already recorded the information on video colors, so color features are unnecessary for video-music correspondence modeling. If we remove remove the influence of tonality by changing keys, we need color features to capture the video colors.

\noindent\textbf{Ablation on Music Generator.}
We further conduct ablation study on our music generator with VMCP. 
To verify the necessity of the decoupled structure,
we test two variants of our model: (1) Video2music: uses the output video features of the fusion encoder to directly generate target music by a Transformer decoder without decoupling the structure of chords, melody, and accompaniment; 
(2) Video2chord2music: generate chords first and then use chords to generate music without decoupling melody and accompaniment. 
As shown in Tab.~\ref{tab:cmt}, 
removing any one or more components of chords, melody, and accompaniment hurts the overall performance of correspondence while having similar music quality. The difference in correspondence and music quality validates that decoupled structure is important for music generation and imposing video controls. 
The improvement of VMCP from Video2music to Video2chord2music shows the effectiveness of decoupling chords, and the improvement from Video2chord2music to the full model V-MusProd shows the effectiveness of decoupling melody and accompaniment. 
We note that Video2music sometimes has better music quality. It can be explained that imposing control over music generation can hurt music quality by adding inductive biases.

\begin{table}
\small
\resizebox{0.45\textwidth}{!}{
    \centering
    \begin{tabular}{lcccccc}
    \toprule
        Methods & SC & PE & PCE & EBR & IOI\\ \midrule
        Real (POP909) & 0.965 & 4.455 & 2.774 & 0.005 & 0.125 \\ \midrule
        CP~\cite{cp} & 0.987 & 3.697 & 2.538 & 0.041 & 0.250 \\
        Music Trans.~\cite{music_transformer} & 0.985 & 3.934 & 2.581 & 0.034 & 0.216 \\
        HAT~\cite{hat} & 0.989 & 3.856 & 2.550 & 0.040 & \textbf{0.139} \\ \midrule
        V-MusProd & \textbf{0.967} & \textbf{4.070} & \textbf{2.774} & \textbf{0.005} & 0.171 \\ \bottomrule

    \end{tabular}
}
\vspace{-2mm}
\caption{
\textbf{Results of unconditional generation on POP909 \cite{pop909}.} For all the metrics, \textbf{closer} to Real is better. 
}
\vspace{-4mm}
\label{tab:pop909}
\end{table}

\vspace{-1mm}
\subsection{Unconditional Generation}
\label{exp:uncond}
\vspace{-1mm}

Our method can be directly used in unconditional music generation. We examine V-MusProd against previous music generation methods: (a) HAT~\cite{hat}: a hierarchical model built on multiple transformer-based levels to enhance the structure of music, achieving state-of-the-art generation quality; (b) CP Transformer~\cite{cp}: transformer-based model using 2D music tokens to compress sequence length; (c) Music Transformer~\cite{music_transformer}: the first transformer-based music generation model with improved relative attention. 

All the above methods are trained on POP909\cite{pop909} dataset. We directly use their publicly available demos for evaluation.
We train our V-MusProd on POP909 without video input, \emph{i.e.} training Chord Transformer without cross attention with video features.
The unconditionally generated results are evaluated by music quality metrics in Sec.~\ref{exp:objective}.
As shown in Tab.~\ref{tab:pop909}, our V-MusProd achieves closer results to POP909 training set than previous methods for most of the metrics. This indicates that unconditional music generation can benefit from our decoupling structure.

\vspace{-1mm}\section{Conclusion}
\label{sec:conclusion}
\vspace{-1mm}
In this paper, we have introduced the SymMV dataset, which contains 1140 videos and corresponding background music with rich annotations.
Based on SymMV, we developed a benchmark model V-MusProd. It decouples music into chords, melody, and accompaniment, then utilizes video-music relations of semantic, color, and motion features to guide the generation process. 
We also introduced the VMCP metric based on video-music CLIP to evaluate video-music correspondence. 
With VMCP and subjective evaluation, we prove that V-MusProd outperforms baseline model CMT in correspondence both qualitatively and quantitatively.

\paragraph{Acknowledgement}
This work was supported in part by the National Key R\&D Program of China under Grant 2022ZD0115502, the National Natural Science Foundation of China under Grant 62122010, and the CCF-DiDi GAIA Collaborative Research Funds for Young Scholars.

{\small
\bibliographystyle{ieee_fullname}
\bibliography{egpaper_final}
}

\clearpage
\appendix

\section{SymMV Dataset}

\begin{figure}
    \centering
    \includegraphics[width=0.8\linewidth]{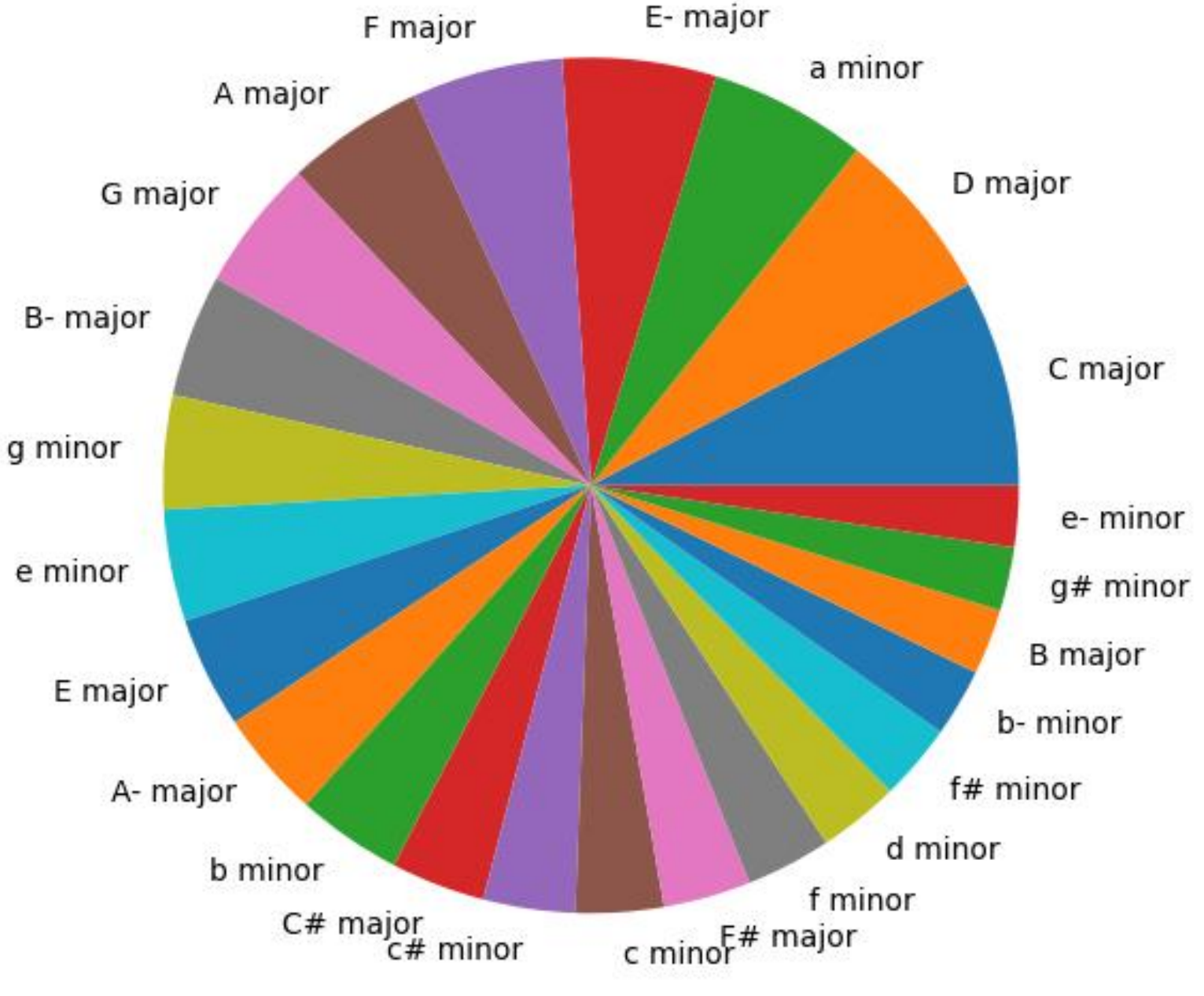}
    \caption{\textbf{Tonality distribution of SymMV.}}
    \label{fig:tonality}
\end{figure}

\subsection{More Collecting Details}
First, we manually search for piano covers and their corresponding music videos, which is cumbersome and time-consuming. 
To enlarge the size of the collected dataset, we search for several Youtube channels of professional piano covers. We download videos with their metadata from these channels and parse the metadata to get the singers and song names. Then we use YoutubeAPI\footnote{https://developers.google.com/youtube/v3} to search for their original music video with keywords in the format of \texttt{\{singer\} + \{song name\} + Official Music Video}. We manually check all collected piano covers and paired videos, where unqualified data like static videos are discarded. Lastly, we utilized ShazamIO\footnote{https://github.com/dotX12/ShazamIO} to recognize the soundtrack from the official music video, which helped us search for the lyrics and genres of the music in SymMV.
 
\subsection{Data Distribution of SymMV}
\begin{figure*}[!t]
    \centering
    \includegraphics[width=0.9\textwidth]{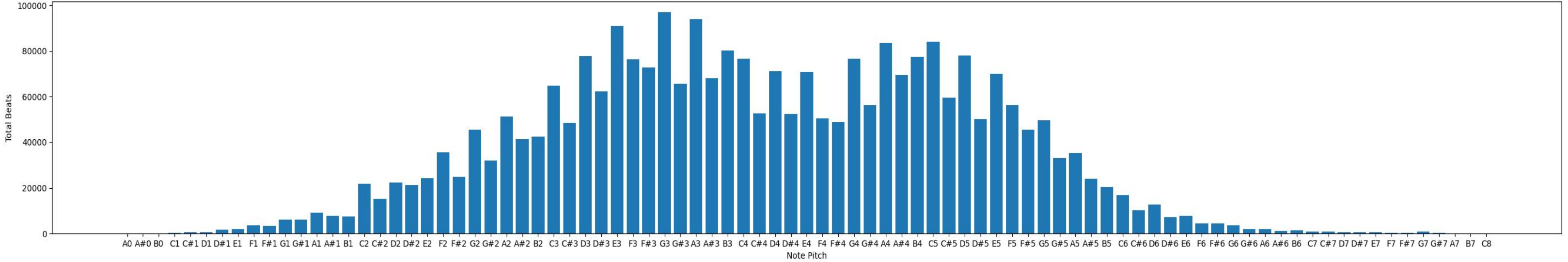}
    \caption{\textbf{Note distribution of SymMV.}}
    \label{fig:note}
\end{figure*}

\begin{figure*}[!t]
    \centering 
    \includegraphics[width=0.9\textwidth]{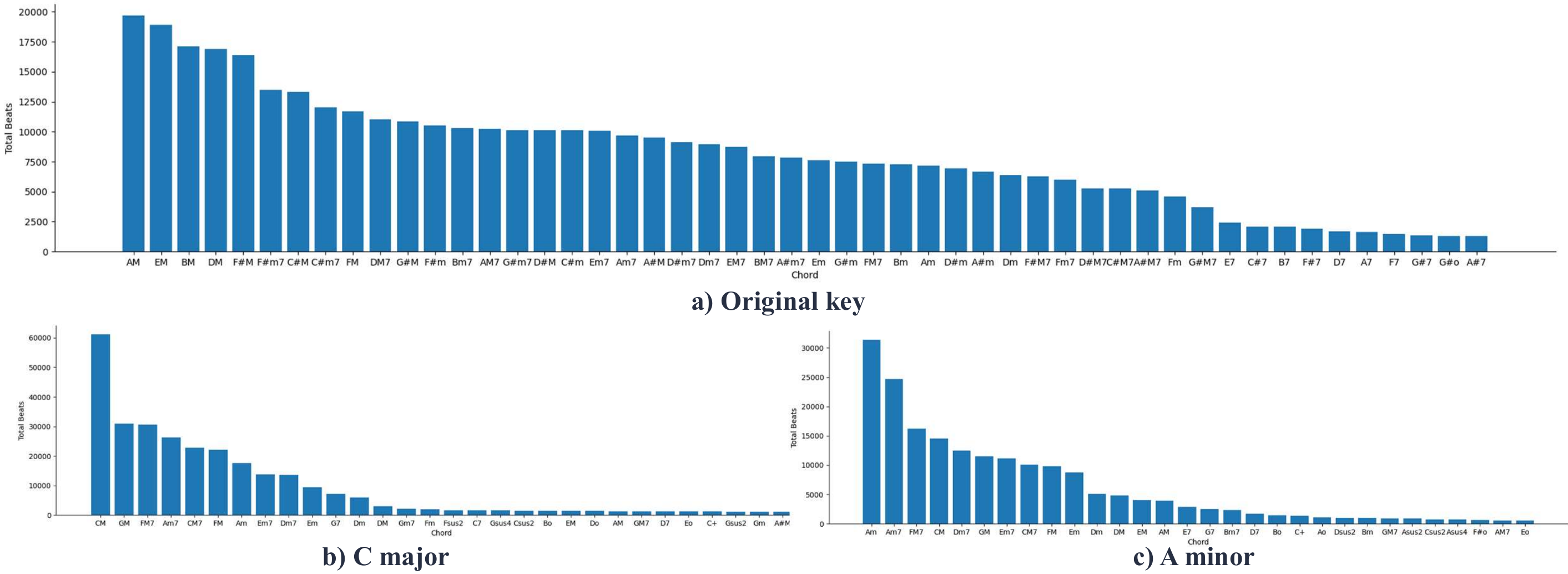}
    \caption{\textbf{Chord distribution of SymMV.} We analyze the chord distribution in both original and modified keys, with the major keys transposed to C major and the minor keys transposed to A minor.}
    \label{fig:chord}
\end{figure*}

We visualize the tonality distribution of SymMV in Fig.~\ref{fig:tonality}. We also compute the statistics of both the total duration beats of each note and each chord in Fig.~\ref{fig:note} and Fig.~\ref{fig:chord}. To eliminate the influence of key on chord distribution for better analysis, we transposed major key music to C major and minor key music to A minor in order to present the chord distribution in the dataset more accurately and intuitively.

\subsection{Analysis of Genres and Chords}

\begin{table}
    \centering
        \begin{tabular}{cc}
            \toprule
            Genre & Count \\
            \midrule
            Pop & 348 \\
            Hip-Hop/Rap & 123\\
            Mandopop & 112 \\
            Alternative & 110\\
            R\&B/Soul & 80\\
            Dance & 63\\
            Soundtrack & 44\\
            Electronic & 31\\
            Singer/Songwriter & 15\\
            K-pop & 12\\
            Rock & 11\\
            Others & -\\
            \bottomrule
        \end{tabular}
\caption{\textbf{Genres of music in SymMV.}}
\label{tab:genres}
\end{table}

\begin{figure}[t]
    \centering
    \includegraphics[width=1.0\linewidth]{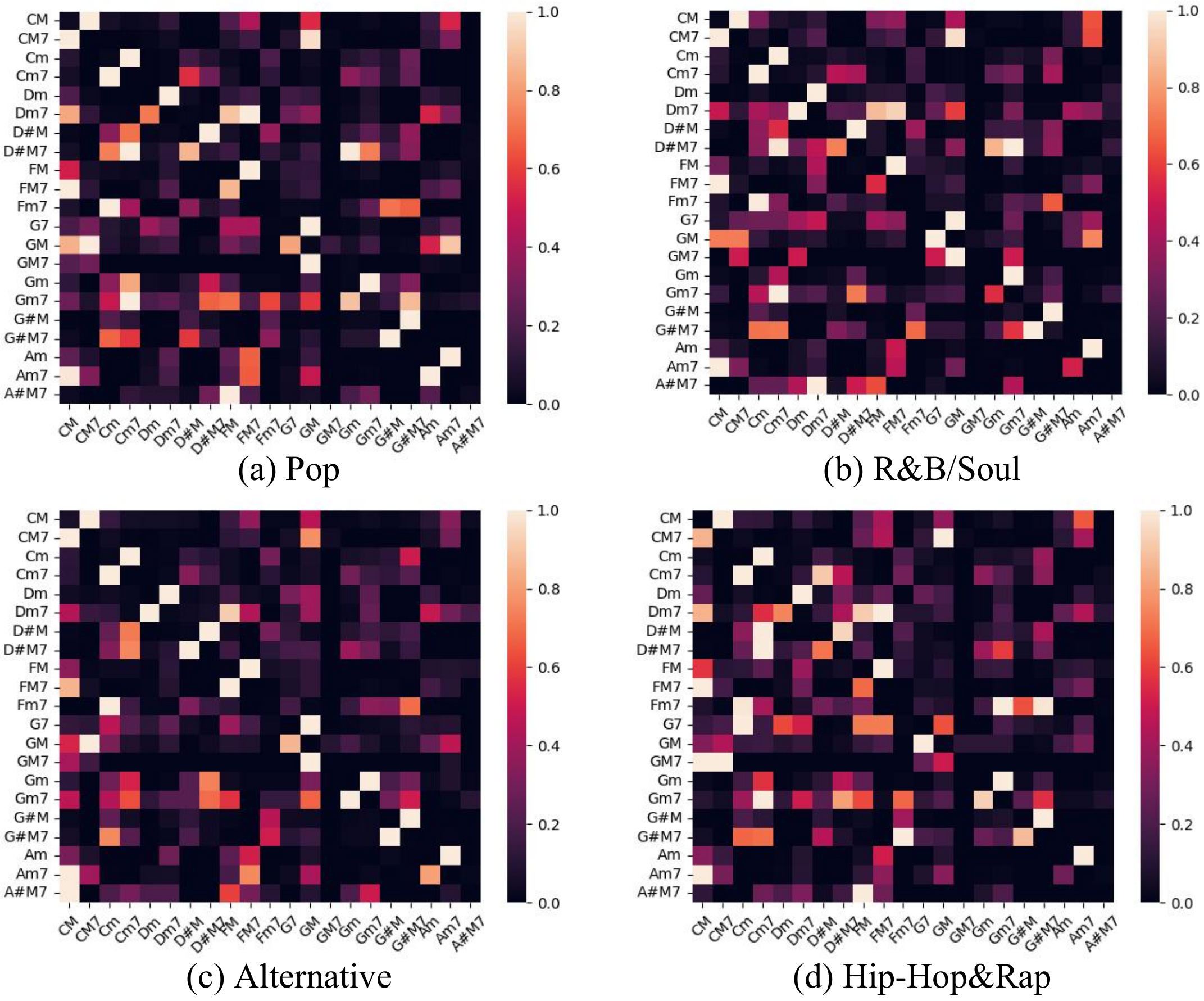}
    \caption{\textbf{Transition matrix for common chords.}}
    \label{fig:transition}
\end{figure}

The number of music in different genres is shown in Tab.~\ref{tab:genres}. In Sec. 3.3, we analyze the chord distribution in different genres. Here we further provide the chord transition matrix for common chords in different genres. As shown in Fig.~\ref{fig:transition}, we observe that the chord transition matrix is asymmetric and sparse, which indicates chord transitions are direction-sensitive and follow the music rules. We ignore the diagonal of the transition matrix in order to highlight the significance of the contrast. 
As Fig.~\ref{fig:transition} shows, the relationship between chords and genres we demonstrate in Sec. 3.3 is further confirmed. \emph{R\&B/Soul} blurs the concept of major and minor keys. For example, chords after CM in other genres are mostly CM7, FM, GM, Am. However, in \emph{R\&B/Soul}, chords after CM can even be Cm, which is unsteady both in major and minor keys.

\begin{table*}[h]
\small
\begin{center}
\begin{tabular}{lllll}
\toprule
Methods & V-MusProd (Ours) & CMT~\cite{cmt} & RhythmicNet\cite{how_does_it_sound} & Foley Music~\cite{foley_music}\\ \midrule

Video Category & General & General & Human activity & Instrument performance \\ \midrule

Videos for Training & Music videos & - & Dance videos & Instrument videos\\ \midrule

Music Category & Generation & Generation & Generation & Reconstruction\\ \midrule

Music Tracks & piano & \begin{tabular}[c]{@{}l@{}}piano, guitar, bass, \\ drum, string\end{tabular} & drum, piano, guitar & \begin{tabular}[c]{@{}l@{}}bass/guitar/piano/violin/...\\(single track)\end{tabular} \\ \midrule

Music Priors & \begin{tabular}[c]{@{}l@{}}Chord, melody, \\ accompaniment\end{tabular}  & Simu-note & - & - \\ \midrule

Music Length & 6min & 3min & 15sec & 6sec \\ \midrule

Video-Music Relations & Semantic, color, motion & Rhythm & Body movements & Body movements \\ \midrule

End-to-end Training & No & Yes & No & Yes \\ 
\bottomrule
\end{tabular}
\end{center}
\caption{\textbf{Comparison with previous video-conditional music generation methods.}}
\label{tab:comparison}
\end{table*}

\section{Comparison with Previous Methods}

We provide comparisons with previous video-conditional music generation methods in Tab.~\ref{tab:comparison}. Further comparison with CMT, the only method to generate music for \emph{general} videos, is as below:

\begin{itemize}[leftmargin=*]
    \item \textbf{Paired Training Data.} We collect and use paired video and music data to achieve learning-based control, while CMT does not use paired data and manually designs empirical rule-based relations trained with music dataset only.
    \item \textbf{Video-music Relations}. V-MusProd considers semantic, color, and rhythmic relations, while CMT only considers rhythmic relations. For rhythmic relations, we keep timing control and add tempo control. We discard density and strength since these two local controls markedly hurt music quality. They are more suitable for multi-track music where some tracks (\emph{e.g.} drum and bass) serve as arrangements for others (\emph{e.g.} piano and guitar).
    \item \textbf{Music Priors.} We adopt music priors of chords, melody, and accompaniment, three important factors for music composition that are consistent with human composition process. CMT only uses simu-note as music theory priors, which is simply for computational modeling and inconsistent with common music composition procedures.
    \item \textbf{Inference Time}. Our inference time is significantly shorter than CMT, yielding a \textbf{20X} inference speed. When running inference for a 3-minute 360p video on one single GPU, CMT takes about 15 minutes to extract video features and 5 minutes to generate MIDI, while ours takes less than 20 seconds to extract features and 40 seconds to generate. For even longer videos of 5 minutes, CMT takes an extremely long time, while our V-MusProd can still generate music within a reasonable time.
\end{itemize}

\section{Implementation Details}
Hyper-parameters of our model are given in Tab.~\ref{tab:hyperparameters}.

\begin{table}[h]
\small
\resizebox{0.5\textwidth}{!}{
    \begin{tabular}{lc}
    \toprule
    (C,M,A) transformer hidden size & 512 \\
    (C,M,A) transformer feed-forward size & 2048 \\
    (C,M,A) transformer attention heads & 8 \\
    (C,M,A) transformer dropout rate & 0.1 \\
    (C,M,A) transformer activation & gelu \\
    (C,M,A) beat resolution & 220 \\
    (C) color encoder layers & 2 \\
    (C) semantic encoder layers & 2 \\
    (C) fusion encoder layers & 2 \\ 
    (M,A) encoder layers & 4 \\
    (M,A) decoder layers & 6 \\
    (C) color feature size & 3072 \\
    (C) semantic feature size & 512 \\
    (M,A) timing encoding size & 256 \\
    
    \midrule
    (C,M,A) initial learning rate & 1e-4 \\
    (C,M,A) optimizer & Adam \\
    (C,M,A) decoder layers & 4 \\
    (C,M) number of epochs & 200 \\
    (A) number of epochs & 400 \\
    (C) batch size & 8 \\
    (M) batch size & 10 \\
    (A) batch size & 3 \\
    
    \midrule
    \begin{tabular}[c]{@{}l@{}}(C) decoder embedding size \\ (barbeat, type, root, quality)\end{tabular} & 128, 32, 128, 64 \\
    \begin{tabular}[c]{@{}l@{}}(M,A) decoder embedding size \\ (barbeat, type, pitch, duration)\end{tabular} & 128, 32, 512, 128 \\
    \begin{tabular}[c]{@{}l@{}}(M) encoder embedding size \\ (barbeat, type, chroma)\end{tabular} & 128, 32, 128 \\
    \begin{tabular}[c]{@{}l@{}}(A) encoder embedding size \\ (barbeat, type, pitch, duration, chroma)\end{tabular} & 128, 32, 512, 128, 128  \\ 
    \bottomrule
    \end{tabular}
}
\caption{\textbf{Hyper-parameters used in our model.} C, M and A denote Chord Transformer, Melody Transformer, and Accompaniment Transformer, respectively.}
\label{tab:hyperparameters}
\end{table}
 
\section{User Study}
\label{supp:subjective}
\begin{figure*}[!t]
    \centering
    \includegraphics[width=\textwidth]{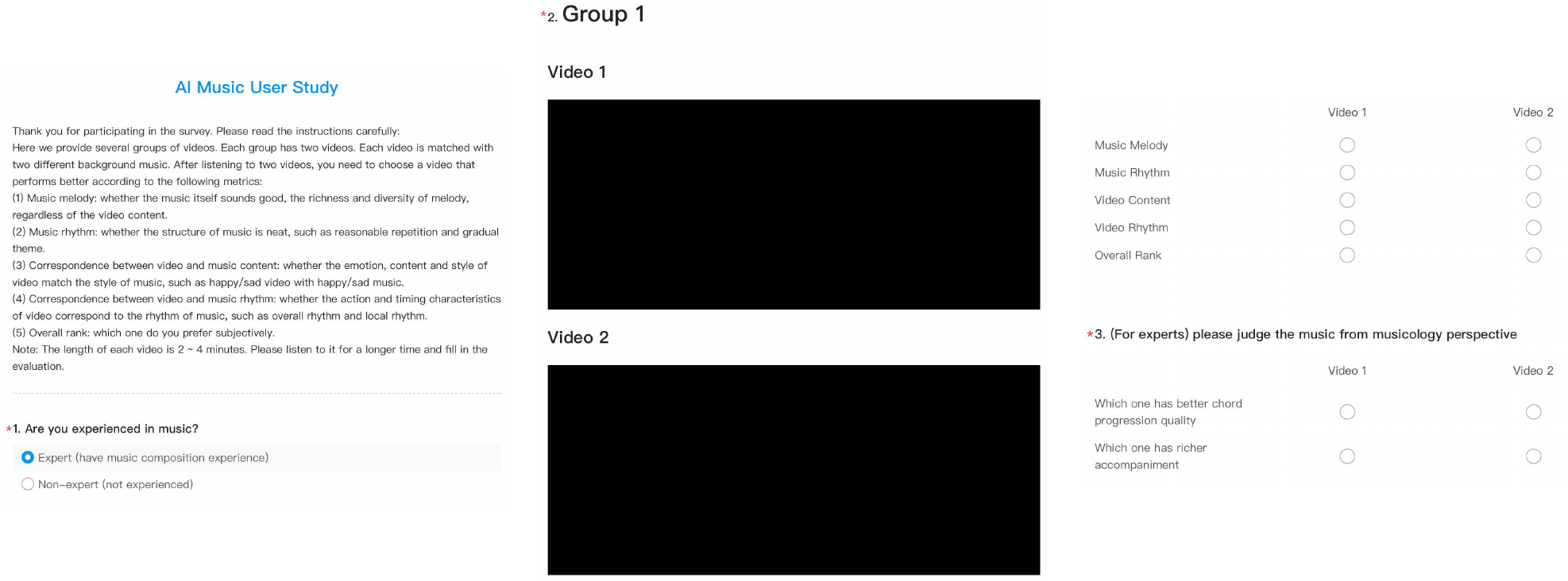}
    \caption{\textbf{Screenshot of the user study questionnaire in subjective evaluation.} We survey professional students in music majors as our experts and regular users as non-experts. The first question serves only to distinguish the experts.}
    \label{fig:userstudy}
\end{figure*}

For the user study in Sec. 6.3, we provide $3$ groups ($6$ videos), each with two pieces of background music by V-MusProd and CMT, respectively. Participants are asked to compare two samples by $5$ (non-expert group) or $7$ metrics (expert group). The questionnaire takes about $20$ minutes to complete. The reward for each participant is \$$5$. In Fig.~\ref{fig:userstudy}, we provide a screenshot of the questionnaire with full text of instructions.

\section{Broader Impacts and Future Directions}

\textbf{Broader Impacts}. Since our dataset is collected from the Internet, it may contain potentially biased information. It needs to be handled carefully to prevent improper usage.

Our method can help video creators generate background music automatically, which saves plenty of time and enhances video quality. A potential concern is that some segments of the generated music may be similar to the training data. Inspections should be conducted carefully on copying from human-written songs that may raise copyright issues. 

As AI-composed music develops in the future, it may compete with human composers, leading to employment-related issues. However, human and AI can also work together to create new styles of music or compose music to fulfill our needs. We believe our exploration of controllable music generation can help to promote the interaction between human artists and AI.

\textbf{Limitations and Future Works.}
Future directions are abundant and worth exploring. Our dataset and method mainly focus on pop music, while other music genres, such as classical music, can further be explored. Since we only generate symbolic music with piano tracks, other instruments (\emph{e.g.} drum, guitar, and string) can still be explored. 
Our method does not model high-level emotion features or repetitive structures of music phrases (verses, chorus, bridges \emph{etc.}) explicitly. 
Besides, the three stages of our method are trained separately instead of end-to-end. 
Our proposed relationships of the color histogram and RGB difference are heuristic and may not be applicable in all situations. 
Since our method serves as a baseline on the SymMV dataset, we expect future works to explore the above directions.

Furthermore, Our V-MusProd can also be adapted to text-conditional music generation with slight modification. For instance, we can use video descriptions and titles to build semantic relations on Chord Transformer and leverage lyrics to control Melody and Accompaniment Transformers with temporal information. 

It is worth noticing that our SymMV dataset also provides lyric annotations to explore relations among text,  videos, and music.

\end{document}